# Transforming Heart Disease Prediction with Advanced Machine Learning Techniques

Sami Ullah, Muhammad Mohsin Khan

City University of Science in Information Technology,
Peshawar 25000, Pakistan
Email: samipak100@gmail.com, mohsinkhancs.ai@gmail.com

*Abstract*— **Heart disease remains the leading cause of mortality globally, necessitating early and accurate detection to improve patient outcomes. This research focuses on the predictive analysis of heart disease using machine learning (ML) techniques, comparing the performance of multiple classifiers to identify the most accurate and least error-prone method. Two datasets from UCI and Kaggle repositories were utilized, each containing 14 attributes related to heart health indicators. Techniques including J48, Naive Bayes, Logistic Regression, Simple Cart, Bagging, Decision Stump, AdaBoost, Artificial Neural Networks, and Support Vector Machine (SVM) were applied. Evaluation metrics such as Mean Absolute Error (MAE), Relative Absolute Error (RAE), accuracy, precision, recall, and F-measure were used for performance comparison. Results revealed that SVM achieved the highest performance on the UCI dataset, while Simple Cart performed best on the Kaggle dataset, offering the highest accuracy and lowest error rates. The research work concludes that ML models, when properly tuned and validated, can significantly assist in the early diagnosis of heart disease, offering critical support for clinical decision-making. Future work may involve hybrid approaches and the use of more recent datasets to further improve prediction accuracy.**

## 1 Introduction

Heart disease remains the foremost cause of mortality worldwide, responsible for a significant proportion of deaths annually, with the prevalence constantly increasing. In Pakistan, heart disease induced mortality is notably high, with projections indicating alarming increases in coming years.[1] Early and accurate prediction of heart disease is paramount to reducing fatality rates by enabling timely interventions [2].

Machine learning (ML) offers robust tools for disease prediction and risk assessment, demonstrating substantial value in healthcare analytics. Recent advancements have focused on comparative evaluations of ML algorithms for Heart Disease Prediction Models (HDPMs),[3] aiming to enhance diagnostic accuracy while minimizing error rates. Existing studies confirm the potential of data-driven models to support clinicians in rapid decision-making, yet inconsistencies in accuracy and reliability persist.

This paper undertakes a systematic comparative analysis of multiple supervised ML frameworks for heart disease prediction, emphasizing metrics such as precision, recall, F-measure, mean absolute error (MAE), [4] relative absolute error (RAE), and overall accuracy. Using two public datasets, the study identifies optimal classification algorithms and assessment criteria, contributing to improved early detection strategies [5].

The remainder of the paper is organized as follows: Section II reviews related work on ML-based heart disease prediction. Section III describes the proposed methodology and datasets. Section IV presents experimental results and comparative analysis. Section V concludes the paper with implications and future work.

Table 1: Factors, types and Symptoms of HDPM

| Category | Name | Description |
|---|---|---|
| Types | Coronary artery disease | A disease in which plaque deposits restrict coronary blood. |
| | | Vessels resulting in decreased blood and oxygen flow to the heart [6]. |
| | Angina pectoris | A medical term in which limited blood supply to the heart causes pain in the chest [7]. |
| | Congestive heart failure | It's a disease where heart fails to deliver enough blood to the body's other organs. Usually called heart failure [8]. |
| | Cardiomyopathy | The type of disease in which weakening or change in the structure of the heart muscle due to insufficient heart pumping [9]. |
| | Congenital heart disease | It describes when the heart starts to beat irregularly because of a problem with its structure or function. It is a type of birth defect that kids are born with. [10]. |
| | Arrhythmias | It is related to a disturbance in the heart beat's rhythmic motion. The heartbeat can be steady, rapid, or erratic [11]. |
| | Myocarditis | Inflammation of the heart muscle that is often caused by a virus, fungus, or bacteria that attack the heart [12]. |
| Factors | Medical conditions | This includes high blood pressure, high cholesterol level, diabetes, and sleep apnea [13]. |
| | Unhealthy Lifestyle | This includes an unhealthy diet, not enough physical activity, unhealthy weight, smoking, alcohol consumption, stress, and drug usage [14]. |
| | Risk factors which are not under the control | This includes sex, age, and family medical history [15]. |
| Symptoms | Chest Discomfort | Feeling of chest heaviness, pressure, or chest pain, arm, or below the breastbone appears when having heart disease [16]. |
| | Heart burning | Feeling of fullness, indigestion, or feel like heartburn [17]. |
| | Nausea | Sweating, vomiting, nausea, or feeling dizzy; severe weakness, anxiety, or asthma; and rapid or inconsistent heartbeats are signs of cardiac disease [18]. |

# 2 Literature Review

Recent advances in machine learning have significantly influenced the field of heart disease prediction and management, [19] facilitating the development of automated, data-driven diagnostic systems. Numerous studies have explored the efficacy of various classification algorithms and preprocessing strategies using benchmark datasets, [20] each reporting distinct strengths and limitations across a range of evaluation metrics. This literature review critically examines the most relevant research, providing a technical synthesis of the approaches, [21] algorithms, and outcomes found in the current state of the art. Particular attention is given to model performance,[22] dataset diversity, and unresolved challenges, highlighting both progress made and areas requiring further investigation [23].

Mohan et al. [24] introduced a method for Heart Disease Prediction and Management Problems (HDPMP) named "An Effective Classification Rule Technique for Heart Disease Prediction." The research employed a UCI repository dataset to experimentally test different classification rule-based methods, such as Decision Table (DTab), JRip, OneR, and PART. The dataset contained fourteen attributes. The performance of such models was gauged via accuracy and error-based measures including Mean Absolute Error (MAE), Root Mean Squared Error (RMSE), Relative Absolute Error (RAE), and Root Relative Squared Error (RRSE). The post analysis results revealed the Decision Table method to be best among the others, with a 73.6% accuracy.
Chaurasia et al. [25] performed an experimental analysis using a dataset collected from UCI that contained 11 features. The J48 Decision Tree (DT), Bagging, and Naive Bayes (NB) algorithms were the main subjects of the analysis. The accuracy, precision, and recall matrix are the foundation of evaluation metrics. The bagging algorithm's accuracy is determined to be 85.03% following the assessment.
Venkatalakshmi et al. [26] performed a comparative analysis of DT and NB on a dataset of 14 characteristics that was acquired from UCI. Metrics used for evaluation are built on accuracy. After the evaluation, it was shown that NB performs better than other methods, with an average score of 85.03%.

Hlaudi Daniel and Mosima Anna [27] employ J48, Bayes Net, NB, SC, and Reduced Error Pruning Tree (REPTREE) to empirically examine a dataset of 14 attributes that was obtained from UCI. Accuracy-based evaluation metrics are used. According to the post-estimate, the prediction accuracy of J48 is 99%.

Dai et al. [28] performed experimental analysis on clinical data-derived dataset using SVM, AdaBoost, logistic regression (LR), and NB. The core element upon which all evaluation metrics stand is accuracy. The evaluation reveals that the prediction accuracy rate for AdaBoost stands at 82%.

Abdar et al. [29] the study evaluated C5.0 DT framework as an appropriate technique for HDPMP through a comparative analysis against SVM, K-nearest-neighbor (KNN), and artificial neural network (ANN). Authorities obtained the dataset comprising fourteen features from UCI. Evaluation metrics start with accuracy measurements. The accuracy assessment confirmed that C5.0 boasts the best results with an accuracy score of 93%.

The research study by UmairShafique and FiazMajeed and HaseebQaiser and IrfanUl Mustafa used UCI data with 14 features to compare machine learning techniques J48, ANN and NB. The two main evaluation metrics focus on accuracy rates and measurement times. The study demonstrates that NB achieved the highest accuracy rate of 82.9% while J48 scored 77.2% accuracy based on evaluation metrics.

Dbritto et al. [30] involved a performance evaluation through algorithm comparison between NB, DT, KNN, and SVM methods on Cleveland and Hungary and Switzerland and Long Beach and Statlog datasets. The evaluation method utilized accuracy which showed that SVM delivered 80% better results than other techniques.

Muhammad Saqlain, Wahid Hussain, Nazar Saqib, and Muazzam Khan [31] have developed the operational framework "Identification of heart failure by using unstructured data of cardiac patients" for HDPMP. The framework applies an algorithm comparison approach between LR and ANN and SVM and RF and DT and NB. The data comes from the Armed Forces Institute of Cardiology (AFIC) for comparative evaluation purposes. The NB classifier produced the best predictions reaching 86% accuracy along with performance ratings from precision and recall and accuracy metrics.

Weng et al. use RF and gradient boosting (GB) together with LR and ANN to improve the accuracy of the predictive classifier through a comparative study [32]. The researchers obtained their 378,256 patient clinical data from different acquisition sources. The evaluation criteria used Area Under the ROC Curve (AUC) metrics to show how ANN achieved the highest predictive accuracy by 3.6% above other approaches.

Keerthanaet al. [33] implemented an algorithm comparison between DT, NB and RF. A total of fourteen features comprise the dataset originating from UCI. When evaluating performance NB showed superior results than other available techniques according to precision and recall and accuracy tests.

Rairikar et al. [34] studied the capabilities of algorithms including RF and DT and NB through testing these models against UCI datasets 14 attributes which contained 270 instances. The assessment criteria use precision and recall and F-measure alongside ROC and PRC curve for evaluation. Quantitative assessment shows that RF surpasses other methodologies by an advantage of 75%.

Nikhil et al. [35] conduct a comparison of algorithms such as J48, LMT, RF, NB, KNN, and SVM.The dataset, which comprises fourteen features, was acquired from UCI. The evaluation metrics were accuracy and time which demonstrated J48 achieved a superior performance with
56.76% accuracy.

Hasan et al. [36] presents an intelligent algorithm known as Comparative Analysis of
Classification Approaches for Heart Disease Prediction. The algorithm depends on an analysis between KNN along with DT (ID3) and NB and LR and RF algorithms. We obtained the data from UCI datasets 14 which contains 303 data points of different attributes. The evaluation metrics consist of Accuracy, Precision, Recall, Sensitivity, Specificity as well as F1-Score. The research findings demonstrate that LR achieves a performance superior to all other techniques by
92.76%.

Ramalingam et al. [37] applied empirical analysis to SVM, KNN, NB, DT, RF and ensemble models. The dataset was obtained by the research team from the University of California Irvine datasets. SVM achieved higher accuracy than other techniques with a measurement of 92.1% accuracy.

Gultepe et al. [38] conducted a comparison of meta-algorithms through an examination of ensembling J48 and ensembling NB using a UCI dataset containing 303 instances with 14 attributes to evaluate hybrid algorithm performance. The basis of all evaluation metrics remains accuracy. The experiment established that J48 ensemble method reached the highest accuracy level of 81.31

Makumba et al.'s [39] used data mining techniques with decision tree and Naive Bayes and K-nearest neighbor algorithms to develop a model for heart disease prognosis support. The implementation platform for this model runs through the application programming interface of Waikato Environment for Knowledge Analysis called WEKA API. The proposed model uses 14 characteristics obtained from UCI data source. The evaluation metrics operate through accuracy-based standards.

Tamara et al. [40] suggest that NB be implemented in order to evaluate the efficacy of algorithms such as J48 and NB in the determination and diagnosis of cardiac disease. The dataset's comparative analysis The studied dataset comes from the UCI dataset where 303 instances are matched with 14 characteristics. The studied results show that NB achieves higher accuracy versus other methods at 83%.

Muppalaneni et al. [41] conducted an analysis of classification methods using RF, SVM, LR, and GB through ROC curve analysis for congestive heart failure diagnostics. Three hundred and three examples from the UCI dataset form the basis of this study while it contains fourteen features. The obtained results demonstrated LR to be the most precise methodology since it achieved 87% accuracy.

Motarwar et al. [42] designed a bootstrap aggregation method for HDPMP and analyzed it using
RF, NB, SVM and Hoeffding Tree and Logistic Model Tree. Data in the UCI dataset contains 303 instances that are supported by 14 diverse attributes. The main evaluation standard derives from accurate measurement. Among all tested techniques RF established the best performance because its accuracy reached 95.08.

Ware et al. [43] performed an experiential study on HDPM detection through experimentation with RF, KNN, SVM, DT, NB and LR classifiers. The UCI dataset provides a dataset with 303 instances along with 14 attributes. Evaluation methods exist to calculate precision along with recall and accuracy. According to his findings SVM demonstrated better performance than other methods by reaching 89.34.

## Literature Table

| Paper | Techniques Used | Evaluation Metric | Results |
|---|---|---|---|
| Mohan et al. [24], 2013 | DTab, JRip, OneR, and Part | Accuracy and error rates (Mean Absolute Error, Root Mean Square Error, Relative Absolute Error, and Root Relative Squared Error) | Highest accuracy achieved by DTab with 73.6%. |
| Chaurasia et al. [25], 2014 | Naive Bayes, J48 | Accuracy, Precision, Recall | Highest accuracy achieved by Bagging with 85.03%. |
| | Decision Tree, Bagging algorithm | | |
| Vekatalakshmi et al. [26], 2014 | Decision Tree, Naive Bayes | Accuracy | Highest accuracy achieved by Naive Bayes with 85.03%. |
| Hlaudi et al. [27], 2014 | J48, Bayes Net, Naive Bayes, Cart, REPTREE | Accuracy | Highest accuracy achieved by J48 with 99%. |
| Dai et al. [28], 2014 | Support Vector Machine, AdaBoost, Logistic Regression, Naive Bayes | Accuracy and error rate | Highest accuracy achieved by AdaBoost with 82%. |
| Abdar et al. [29], 2015 | C5.0, Support Vector Machine, kNearest Neighbors, Artificial Neural Network | Accuracy | Highest accuracy achieved by C5.0 with 93%. |
| Shafique et al. [44], 2015 | Decision Tree, Artificial Neural Network, Naive Bayes | Accuracy, Receiver Operating Characteristic curve, Time | Highest accuracy achieved by Naive Bayes with 82.9%. |
| Dbritto et al. [30], 2016 | Naive Bayes, Decision Tree, k-Nearest Neighbors, Support Vector Machine | Accuracy | Highest accuracy achieved by Support Vector Machine with 80%. |
| Saqlain et al. [31], 2016 | Logistic Regression, Artificial Neural Network, Support Vector Machine, Random Forest, Decision Tree, Naive Bayes | Accuracy, Precision, Recall | Highest accuracy achieved by Naive Bayes with 86%. |
| Weng et al. [32], 2017 | Random Forest, Logistic Regression, Gradient Boosting, Artificial Neural Network | Area Under Curve | Improved accuracy by Artificial Neural Network with 3.6%. |
| Keerthana et al. [33], 2017 | Naive Bayes, Decision Tree (J48), Random Forest | Accuracy, Precision, Recall | Naive Bayes performs better. |
| Rairikar et al. [34], 2018 | Random Forest, Decision Tree, Naive Bayes | Precision, Recall, F-Measure, Receiver Operating Characteristic curve, Precision Recall curve | Highest accuracy achieved by Random Forest with 81%. |
| Nikhil et al. [35], 2018 | k-Nearest Neighbors, Support Vector Machine, J48, Logistic Model Tree, Random Forest, Naive Bayes | Accuracy, Time | J48 with 56.76% accuracy. |
| Hasn et al. [36], 2018 | k-Nearest Neighbors, Decision Tree, Naive Bayes, Logistic Regression, Random Forest | Precision, Receiver Operating Characteristic, F-Measure, Precision Recall curve, Accuracy, Sensitivity, Specificity | Highest accuracy achieved by Logistic Regression with 92.76%. |
| Ramalingam et al. [37], 2018 | Support Vector Machine, k-Nearest Neighbors, Naive Bayes, Decision Tree, Random Forest, ensemble models | Accuracy | Highest accuracy achieved by Support Vector Machine with 92.1%. |
| Gultepe et al. [38], 2019 | Ensembling J48, Ensembling Naive Bayes | Precision, Recall, Accuracy | Highest accuracy achieved by assembling J48 with 81.31%. |
| Makumba et al. [39], 2019 | Decision Tree, Naive Bayes, k-Nearest Neighbors | Confusion Matrix | Highest accuracy achieved by Decision Tree with 97.07%. |
| Tamara et al. [40], 2019 | J84, Naive Bayes | Accuracy, Time | Highest accuracy achieved by ensembling Naive Bayes with 83%. |
| Muppalaneni et al. [41], 2019 | Random Forest, Support Vector Machine, Logistic Regression, Gradient Boosting | Receiver Operating Characteristic curve | Highest accuracy achieved by Logistic Regression with 87%. |
| Motarwar et al. [42], 2020 | Random Forest, Naive Bayes, Support Vector Machine, Hoeffding Tree, Logistic Model Tree | Accuracy | Highest accuracy achieved by Random Forest with 95.08%. |

| Ware et al. [43], 2020 | Support Vector Machine, Random Forest, k-Nearest Neighbors, Decision Tree, Naive Bayes, Logistic Regression | Precision, Recall, F-Measure | Highest accuracy achieved by Support Vector Machine with 89.34%. |
|---|---|---|---|
| Barik et al. [45], 2020 | Naive Bayes, Decision Tree, Logistic Regression, Random Forest | Precision, Recall, F-Measure | Highest accuracy achieved by Random Forest with 90.16%. |

# 3 Methodology

A compilation of two distinct HDPMDs, the UCI dataset and the Kaggle dataset, marks the beginning of the detailed methodology. Following the dataset's accumulation, classification techniques are implemented to enhance its accuracy and reduce its error rate. To achieve this, the dataset is initially trained using a 10-fold cross-validation (CV) with a variety of techniques, such as J48, NB, LR, SC, Bagging, DS, AdaBoost, ANN, REPT, and SVM. Subsequently, each technique is assigned a prediction task. After the prediction, an analysis of comparisons was conducted among all of the aforementioned techniques to determine which one has a lower error rate and a higher level of accuracy. Figure ?? illustrates the comprehensive methodology for HDPMP.

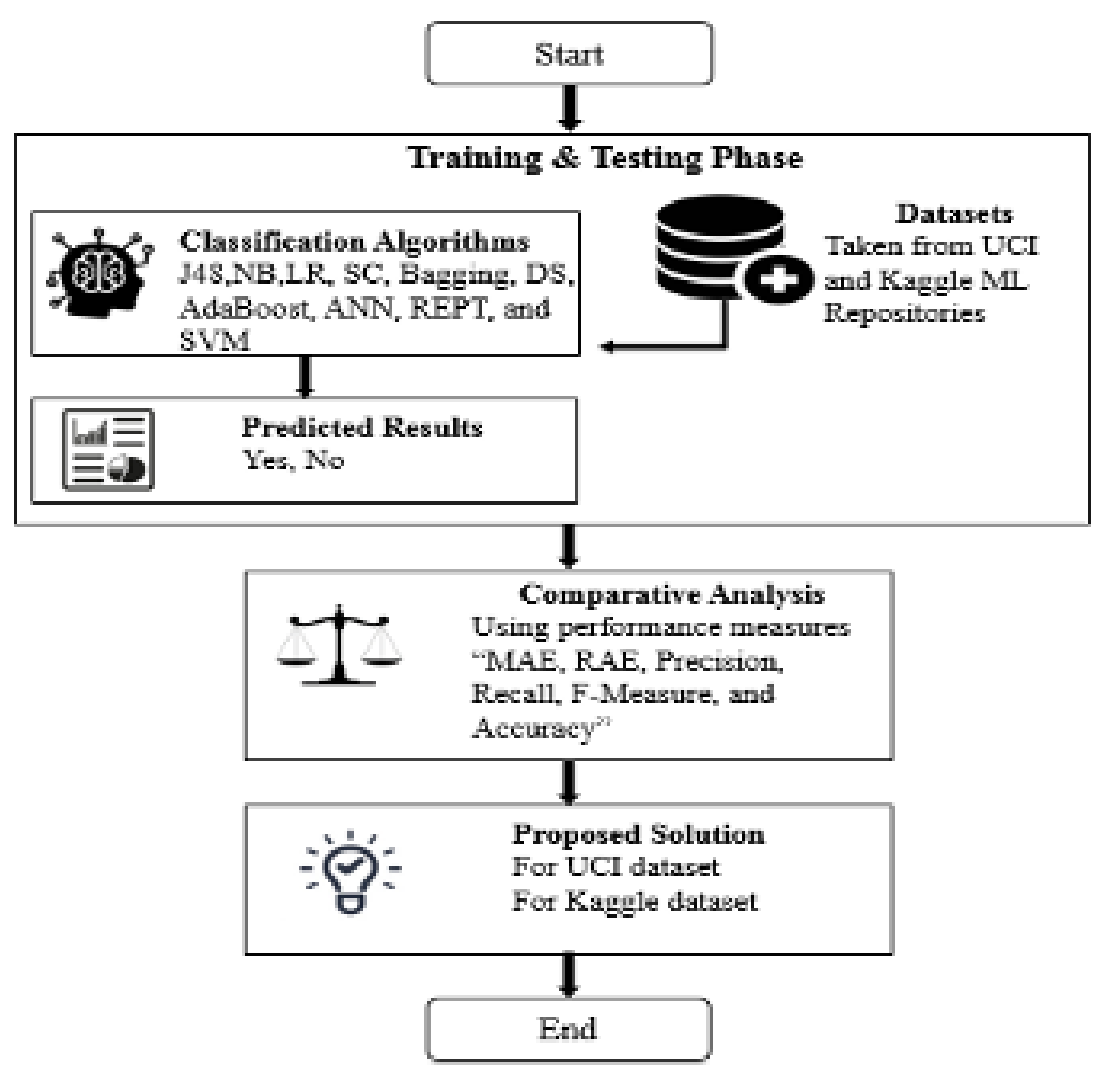

Figure 1: Comprehensive methodology for HDPMP

## 3.1 Datasets

The research examines machine learning classification methods by collecting data from both UCI and Kaggle repositories. The UCI repository consists of 303 instances which are spread across 14 features while the Kaggle ML repository distributes its 1025 instances into 14 attributes. The two databases have 13 independent variables that serve as their attributes. The information contains patient age portrayed as an interval value and gender and three categories of chest pain as well as blood pressure (between 90–250) and cholesterol level (200–250 upwards) and blood sugar (nominal at 120–120) with the last two features being exercise cardiac rate and maximum cardiac rate. For further details, refer to the UCI repository dataset.

**Table 3.1 : List of Attributes**

| S No | Variable | Description | Measurement Scale |
|---|---|---|---|
| 1 | Age | Age of patient | Interval |
| 2 | Sex | Sex of patient | Nominal |
| 3 | Cp | Chest pain type | Nominal |
| 4 | Trestbps | Resting blood pressure | Interval |
| 5 | Chol | Cholesterol level | Interval |
| 6 | Fbs | Fasting blood sugar | Nominal |
| 7 | restecg | Resting electrographic results | Ratio |
| 8 | thalach | Maximum exercise heart rate achieved (bpm) | Interval |
| 9 | Exang | Exercise induced angina | Nominal |
| 10 | Oldpeak | ST depression induced by exercise relative to rest | Interval |

| 11 | Slope | The slope of the peak exercise ST segment | Nominal |
|---|---|---|---|
| 12 | Ca | Number of major vessels colored by fluoroscopy | Interval |
| 13 | Thal | Normal, fixed defect, reversible defect | Nominal |
| 14 | Target | Absence, Presence | Nominal |

**Table 3.2: Attributes and Values Range**

| SN | Variables | Range |
|---|---|---|
| 1 | Age | In Years |
| 2 | Sex | 1: Male; value 0: Female |
| 3 | Cp | value 1: typical type 1 angina; value 2: typical type angina; value 3: non-angina pain; value 4: asymptomatic |
| 4 | Trestbps | mm Hg |
| 5 | Chol | 200-250 or higher mg/dL |
| 6 | Fbs | (value 1: ¿120 mg/dl; value 0: ¡120 mg/dl) |
| 7 | Restecg | Value 0: Normal ; Value 1: ST-wave abnormality ; Value 2: probable left ventricular hypertrophy |
| 8 | Thalach | Beats per minute (bmp) |
| 9 | Exang | value 1: yes ; value 0: no |
| 10 | Oldpeak | 1-3 |
| 11 | Slope | Value 1: unsloping; Value 2: flat; Value 3: down sloping |
| 12 | Ca | value 0-3 |
| 13 | Thal | value 3: normal; value 6: fixed defect; value 7: reversible defect |
| 14 | Target | 0 or 1 |

To evaluate the effectiveness of heart disease prediction models, all experiments in this study utilized a rigorous 10-fold cross-validation protocol. This process starts by dividing each dataset into ten equal parts. For every round, the model is trained on nine parts and validated on the remaining subset. Rotating the validation set in this manner ensures that each record is tested for generalization, greatly reducing bias and providing a robust overview of each model's performance. Such repeated validation captures both model accuracy and potential weaknesses, making it a standard approach for benchmarking machine learning algorithms.

A wide range of classification algorithms were deployed to explore which approaches deliver optimal results on complex health datasets. Tree-based techniques like J48 use gain ratio and pruning strategies to construct reliable decision boundaries while mitigating overfitting. AdaBoost and Bagging aggregate multiple weak learners, each trained on random data subsets, to produce ensemble models that are both accurate and resilient to data noise. When it comes to neural network approaches, models such as the artificial neural network implement multi-layer architectures that simulate human neuron interactions, allowing the system to learn complex, nonlinear patterns. Reduced Error Pruning Trees speed up decision tree learning by trimming branches that do not improve prediction accuracy, and support vector machines search for optimal separating hyperplanes in high-dimensional feature spaces, maximizing class separation and adapting to both linear and nonlinear patterns.

Additional methods, including logistic regression, naive Bayes, simple CART, decision stump, and others, target a balance between model interpretability, speed, and predictive power. Logistic regression employs mathematical optimization updating coefficients through gradient descent to map input features to risk probabilities. Naive Bayes algorithms leverage statistical independence and likelihoods for fast classification, while CART and decision stump perform binary splits based on entropy or threshold measures, yielding interpretable tree structures that can highlight critical attributes.

Each classifier was not merely scored on a single metric. Instead, a comprehensive assessment framework was implemented including mean absolute error, relative absolute error, accuracy, precision, recall, and F-measure. These indicators provided nuanced insights: for example, mean absolute error illustrates the deviation between predicted and actual values, while precision and recall reveal a model's ability to correctly identify the positive class without excessive false alarms or missed detections.

The computational experiments took place in a robust Java environment using Eclipse as the integrated development platform and the Weka software library for machine learning routines. This combination allowed researchers to efficiently organize data preprocessing, model training, and batch evaluation for each algorithm under study. By conducting all analyses within a unified pipeline, consistency was ensured across model comparisons, and results could be directly attributed to methodological differences rather than implementation constraints.

In summary, this methodology offers a thorough and balanced evaluation of various algorithms for heart disease prediction. By employing cross-validation, measuring multiple performance metrics, and utilizing proven computational tools, the study addresses both the reliability and practical limitations of machine learning models, aiming to recommend approaches that maximize predictive accuracy and clinical value.

# 4 Experimental Results

This chapter presents the experimental findings comparing SC and SVM approaches, including outcomes from the baseline method. The discussion begins with results from a range of applied models, and proceeds to a focused evaluation of SC and SVM. Two distinct HDPMD algorithms were utilized for result collection. Key evaluation metrics include mean absolute error (MAE), relative absolute error (RAE), accuracy, precision, recall, and f-measure. This chapter provides results and relevant comparisons using these metrics.

The analysis conducted in this section evaluates the performance of multiple classification algorithms. Six assessment metrics are used to analyze information from two datasets. Ten classifiers are included: J48, Naive Bayes, logistic regression, simple CART, Bagging, decision stump, AdaBoost, artificial neural network, reduced error pruning tree, and support vector machine. These classifiers are applied to HDPMD datasets from UCI and Kaggle using 10-fold stratified cross-validation. Evaluation covers RAE, MAE, correct and incorrect prediction counts, accuracy, precision, recall, and f-measure, allowing an effective comparison to identify which HDPM method performs best.

The UCI dataset analysis displays algorithm performance as shown in Table ??, reporting correct classified instances (CCI) and incorrect classified instances (ICI) for each method. The data provide both accurate and error counts per algorithm and instance. Notably, SVM achieves the highest accuracy, correctly classifying 253 instances and misclassifying 50, outperforming other methods in Table.

**Table 4.1 : Correct and Incorrect Instances on UCI Dataset**

| S No | Technique | CI | CI (%) | ICI | ICI (%) |
|---|---|---|---|---|---|
| 1 | SC | 248 | 81.5% | 55 | 18.15% |
| 2 | J48 | 238 | 78.55% | 65 | 21.45% |
| 3 | ANN | 236 | 77.89% | 67 | 22.11% |
| 4 | Bagging | 249 | 82.18% | 54 | 17.82% |
| 5 | REPTree | 240 | 79% | 63 | 20.79% |
| 6 | LR | 249 | 82.1% | 54 | 17.82% |
| 7 | AdaBoost | 247 | 81.51% | 56 | 18.48% |
| 8 | NB | 251 | 82.8% | 52 | 17.16% |
| 9 | DS | 225 | 74.26% | 78 | 25.74% |
| 10 | SVM | 253 | 83.49% | 50 | 16.50% |

The error rates for the employed techniques, which include relative absolute error (RAE) and mean absolute error (MAE), are presented in Table. It is evident from Table that SVM yields superior error rates, with a RAE value of 33.26% and an MAE value of 0.16.

**Table 4.2 : Experimental Error Rates of Employed Techniques on UCI dataset**

| S.No | Techniques | RAE (%) | MAE |
|---|---|---|---|
| 1 | SC | 53.70 | 0.26 |
| 2 | J48 | 50.28 | 0.24 |
| 3 | ANN | 43.79 | 0.21 |
| 4 | Bagging | 56.30 | 0.27 |
| 5 | REP Tree | 57.00 | 0.28 |
| 6 | LR | 47.00 | 0.23 |
| 7 | Ada Boost | 46.38 | 0.23 |
| 8 | NB | 41.60 | 0.20 |

| 9 | DS | 75.24 | 0.37 |
|---|---|---|---|
| 10 | **SVM** | **33.26** | **0.16** |

The results show that SVM achieved superior performance than alternative methods on UCI dataset with lower error rates as presented in Table 4.3.

**Table 4.3 : Difference in Error Rate on UCI dataset**

| S No | Techniques | Diff in RAE | Diff in MAE |
|---|---|---|---|
| 1 | SVM with SC | 20.44 | 0.1 |
| 2 | SVM with J48 | 17.02 | 0.08 |
| 3 | SVM with ANN | 10.53 | 0.05 |
| 4 | SVM with Bagging | 23.04 | 0.11 |
| 5 | SVM with REPTree | 23.74 | 0.12 |
| 6 | SVM with LR | 13.74 | 0.07 |
| 7 | SVM with AdaBoost | 13.12 | 0.07 |
| 8 | SVM with NB | 8.34 | 0.04 |
| 9 | SVM with DS | 41.98 | 0.21 |

Some metric should be established to predict the validity of algorithms during evaluation. Accuracy is of the utmost importance in order to determine the extent to which it works accurately Table 4.4 presents a detailed analysis of precision, recall, F-measure, and accuracy numbers for the classifiers. The results indicate that SVM provides superior results than other implemented algorithms. The performance analysis for precision and recall as well as F-measure appears in Figure 4.1 while Figure 4.2 presents accuracy metrics.

**Table 4.4 : Accuracy of Employed Classifiers on UCI dataset**

| Technique | Precision | Recall | F-Measure | Accuracy (%) |
|---|---|---|---|---|
| SC | 0.823 | 0.818 | 0.816 | 81.5 |
| J48 | 0.785 | 0.785 | 0.785 | 78.55 |
| ANN | 0.77 | 0.77 | 0.77 | 77.89 |
| Bagging | 0.82 | 0.82 | 0.82 | 82.18 |
| REPTree | 0.79 | 0.79 | 0.78 | 79.00 |
| LR | 0.82 | 0.82 | 0.82 | 82.10 |
| AdaBoost | 0.81 | 0.81 | 0.81 | 81.51 |
| NB | 0.83 | 0.82 | 0.82 | 82.80 |
| DS | 0.74 | 0.74 | 0.74 | 74.26 |
| SVM | 0.84 | 0.83 | 0.83 | 83.49 |
| | | | | |

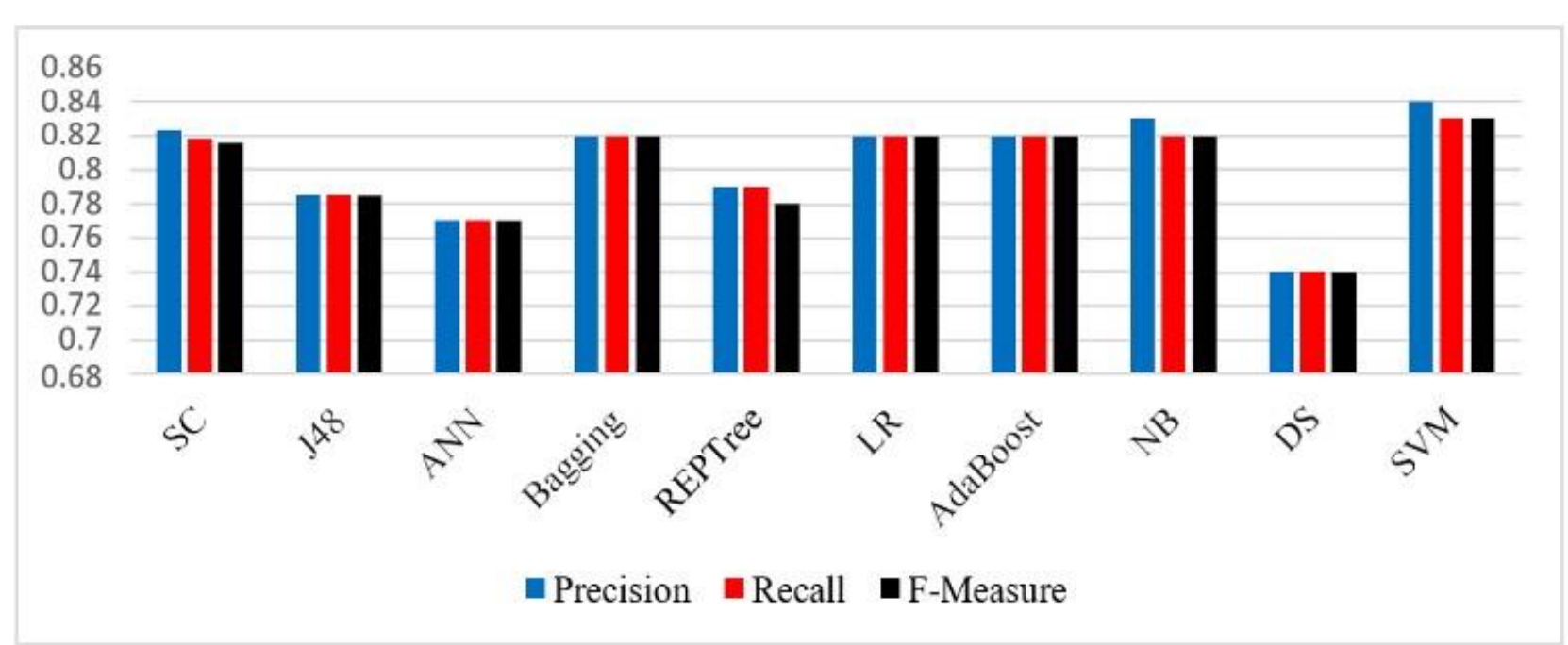

Figure 2: Precision, Recall and F-measure Analysis on UCI dataset

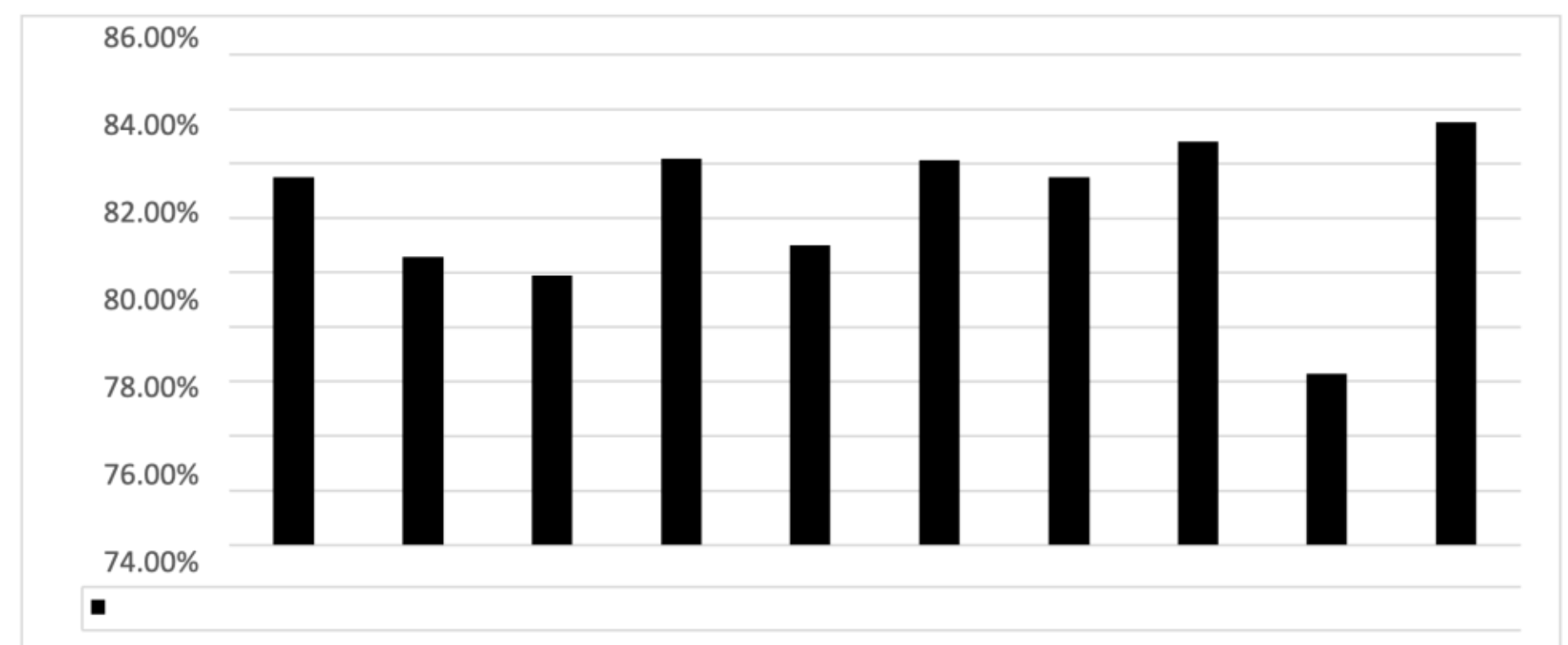

Figure 3: Accuracy Achieved vie each Classifier on UCI dataset

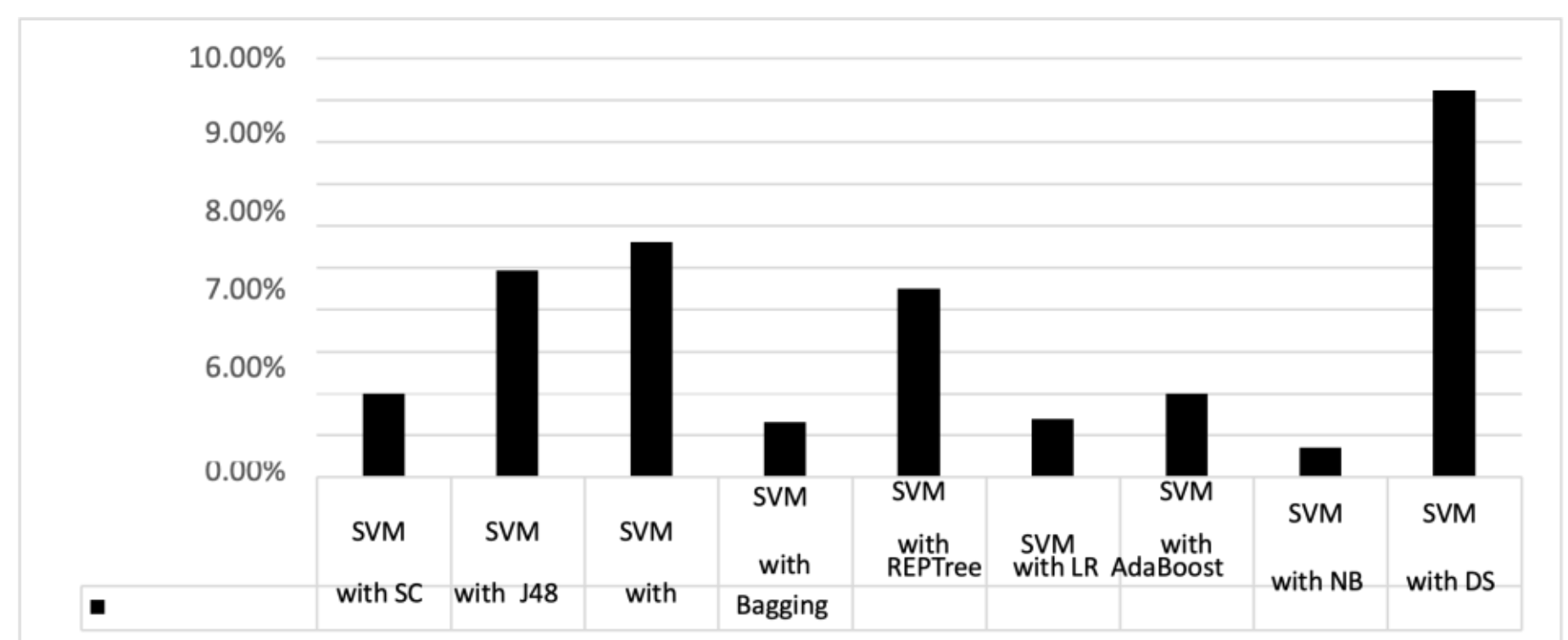

Figure 4: Accuracy Difference between SVM and Employed Classifiers:

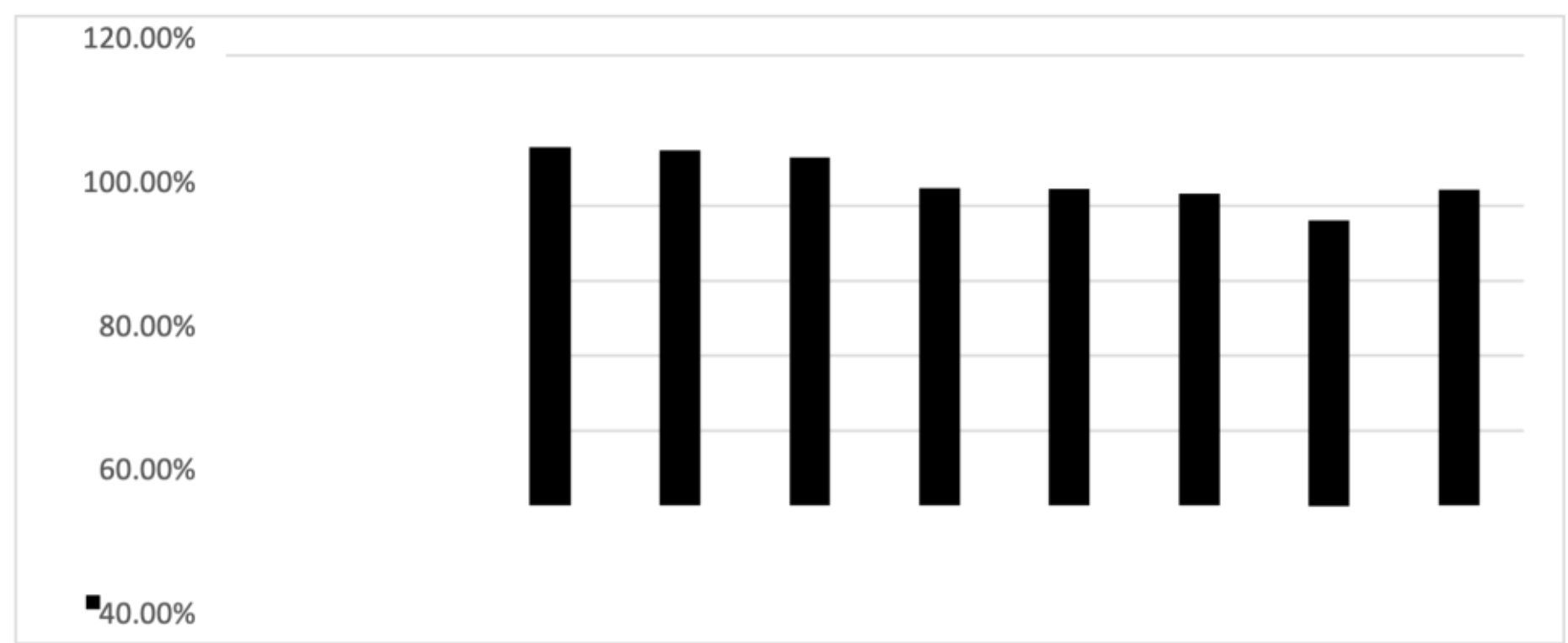

Figure 5: Accuracy Achieved vie each Classifier:

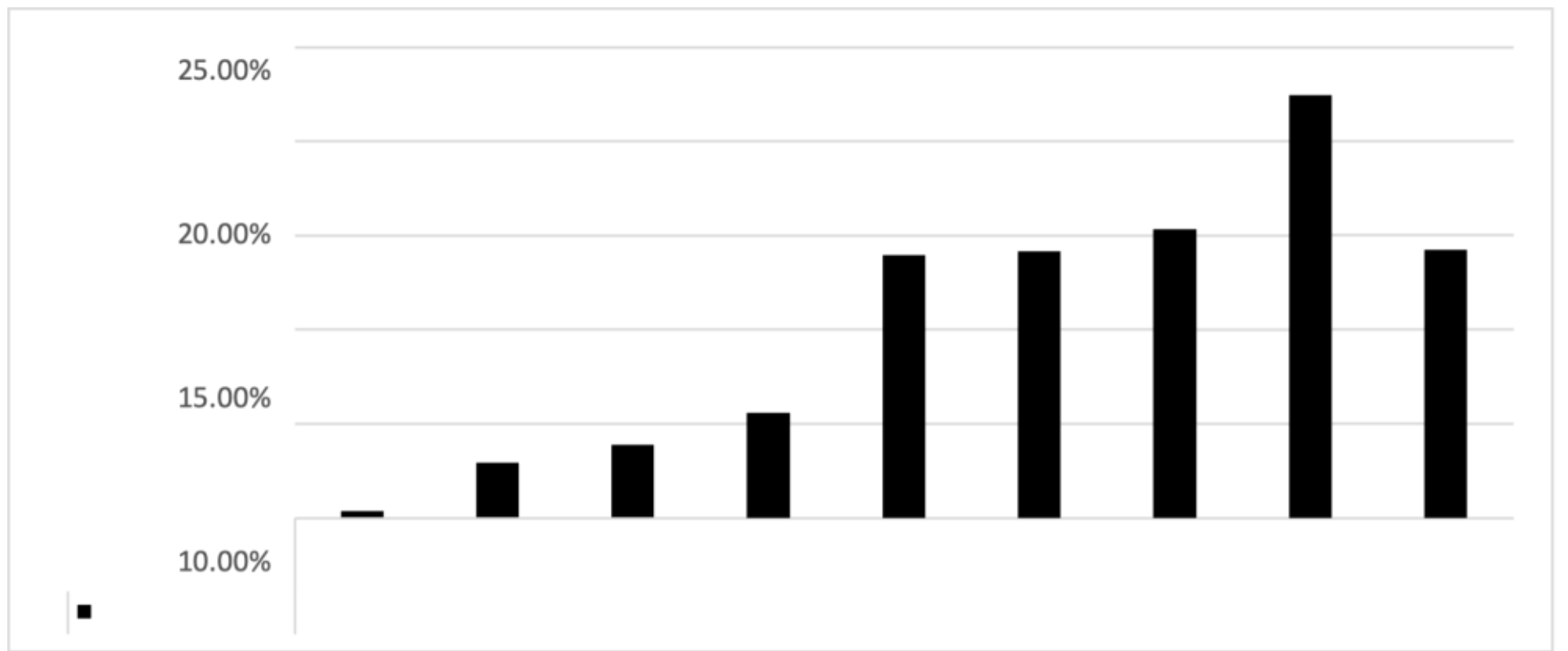

Figure 6: Accuracy Difference between SC and other Employed Classifiers:

This research evaluates ten different ML classification algorithms across two distinctive HDPMDs to analyze their performance. The algorithms originated from the UCI Ml repository in addition to the Kaggle repository. Each dataset contains unique features which lead to different instance numbers as demonstrated through their evaluation process. The datasets present distinctive ratios between effective and ineffective patient records. A comparison of results in Table 4.9 illustrates how improved classification mechanisms achieved better performance for both datasets according to all chosen evaluation metrics. Both datasets confirm that maximizing accuracy is possible along with reducing error rates. The UCI dataset demonstrates superior performance from SVM for accuracy alongside f-measure and precision and recall and error rates. The SC dataset from the Kaggle repository achieves superior outcomes that drive increased recall accuracy precision and F-measure performance while simultaneously decreasing error rates particularly MAE and RAE.

# 5 Conclusion

evaluation metrics. Specifically, two algorithms—SC for the Kaggle datasets and SVM for the UCI datasets—were examined in detail. For both the UCI and Kaggle datasets, evaluation relied on the calculation of MAE and RAE, along with accuracy, precision, recall, and f-measure to benchmark performance. Results from comparative analysis of these approaches and newly proposed models serve as the foundation for evaluating improvements in predictive capability.

Optimizing accuracy and reducing error rates remain primary objectives for future research in HDPMP. Progress in this area may be achieved by embracing updated datasets and more sophisticated algorithms, which can raise the level of accuracy further. Additionally, fusing the strengths of both SC and SVM into hybrid approaches represents a promising direction for boosting both efficiency and effectiveness.

Consideration of potential threats to validity is essential for interpreting the findings of this research. Regarding internal validity, the study draws upon established assessment criteria that have been widely adopted in prior investigations. These criteria fall into two categories: those that are designed to measure error rates and those that assess accuracy. The reliability of the accuracy metrics may be affected if new standards supplant those already in use, and if research methodologies are updated with contemporary procedures.

External validity is influenced by the use of datasets from the UCI ML repository and Kaggle repository. The generalizability of the outcomes could be compromised if the developed techniques are evaluated using medical organization data rather than the selected datasets, or if other datasets are substituted, potentially impacting the accuracy rates. Specific data requirements may challenge the adaptability of the proposed models to broader prediction scenarios.

Construct validity is supported by employing multiple machine learning approaches from widely recognized The literature review demonstrates that various research projects have proposed methods for HDPMP, yet the goal of substantially lowering error rates and increasing accuracy remains elusive. Despite extensive investigation, the primary causes of persistent predictive limitations have not been fully determined, and solutions to overcome these challenges have yet to be realized. The present study focuses on advancing HDPMP accuracy by refining repositories, applying diverse evaluation methods to analyze heart disease datasets. The advanced characteristics of the selected techniques underlie their superiority compared to prior work, though the possibility remains that newer approaches may surpass those presented here. Furthermore, continuous improvement may result from refinements in model training and testing protocols, such as adjusting dataset splits or varying the number of cross-validation folds. Updated evaluation standards also have the potential to enhance results beyond the benchmarks attained in this study.